\documentclass[11pt]{article}

\usepackage[preprint]{acl}

\usepackage{times}
\usepackage{latexsym}

\usepackage[T1]{fontenc}

\usepackage[utf8]{inputenc}

\usepackage{microtype}
\usepackage{amsmath}
\usepackage{multicol}
\usepackage{multirow}
\usepackage{cleveref}
\usepackage{booktabs}
\usepackage{float}

\usepackage{inconsolata}

\usepackage{graphicx}

\title{RB-FT: Rationale-Bootstrapped Fine-Tuning for Video Classification}

\newcommand{\myparagraph}[1]{\smallskip\noindent\textbf{#1}}

\author{
\normalfont
Meilong Xu$^{1}$,
Di Fu$^{2}$\thanks{Corresponding author.},
Jiaxing Zhang$^{2}$,
Gong Yu$^{2}$,
Jiayu Zheng$^{2}$,
\\
Xiaoling Hu$^{3}$,
Dongdi Zhao$^{2}$,
Feiyang Li$^{2}$,
Chao Chen$^{1}$,
Yong Cao$^{2}$
\\[0.5em]
$^{1}$Stony Brook University, Stony Brook, NY, USA \\
$^{2}$ByteDance Inc., Seattle, WA, USA / Sydney, Australia \\
$^{3}$Harvard Medical School, Boston, MA, USA \\
\texttt{meixu@cs.stonybrook.edu}, \texttt{fudi.01@bytedance.com}
}

\begin{document}
\maketitle
\begin{abstract}
Vision Language Models (VLMs) are becoming increasingly integral to multimedia understanding; however, they often struggle with domain-specific video classification tasks, particularly in cases with limited data. This stems from a critical \textit{rationale gap}, where sparse domain data is insufficient to bridge the semantic distance between complex spatio-temporal content and abstract classification labels. We propose a two-stage self-improvement paradigm to bridge this gap without new annotations. First, we prompt the VLMs to generate detailed textual rationales for each video, compelling them to articulate the domain-specific logic. The VLM is then fine-tuned on these self-generated rationales, utilizing this intermediate supervision to align its representations with the nuances of the target domain. Second, conventional supervised fine-tuning (SFT) is performed on the task labels, achieving markedly higher effectiveness as a result of the model's pre-acquired domain reasoning. Extensive experiments on diverse datasets demonstrate that our method significantly outperforms direct SFT, validating self-generated rationale as an effective, annotation-efficient paradigm for adapting VLMs to domain-specific video analysis.
\end{abstract}

\section{Introduction}
\label{sec:intro}
Automated video analysis is increasingly vital in modern industrial settings, offering substantial gains in efficiency, safety, and quality control~\citep{tang2025multi}. In manufacturing, for instance, vision systems monitor assembly lines to detect subtle product defects with superhuman consistency~\citep{he2023videopro}, optimize workflows~\citep{ravindran2023internet}, and enforce safety protocols by ensuring compliance with protective equipment standards~\citep{saurez2023utility}. Despite their importance, these applications present a significant challenge for modern computer vision models~\citep{shaar2025adapting}. The video data from specialized industrial environments, which are characterized by repetitive processes and minute visual anomalies, differ drastically from the general, human-centric video datasets (e.g., Kinetics~\citep{kay2017kinetics}) used to train prominent models, creating a substantial ``domain gap'' that hinders out-of-the-box performance.

The current state-of-the-art in visual analysis~\citep{luo2025videoautoarena, ke2025explain, hong2024cogvlm2, ruthardt2024better} is led by Large Vision-Language Models (LVLMs), which are pre-trained on vast internet-scale datasets of image-text pairs to learn a rich, generalizable understanding of the world~\citep{radford2021learning, li2024multimodal, zhou2024vicor, xu2024lvlm}. Their ability to perform zero-shot classification by aligning visual information with natural language prompts makes them incredibly powerful~\citep{li2025survey, spinaci2025benchmarking, baia2025zero, singh2024learn, lavoie2024modeling}. However, this reliance on general-domain knowledge becomes a critical weakness when deploying these models in specialized industrial contexts. The visual and linguistic chasm between web data and industrial video leads to a sharp decline in performance~\citep{abdullah2025isafetybench, radsch2025bridging, fang2024mmbench, li2024videoeval, parashar2024neglected, jiang2024effectiveness}. An LVLM's understanding of an abstract label like ``defective'' is built from diverse web images of torn clothes or dented cans. When faced with a hairline fracture on a turbine blade, the model fails not just because the visual features are unfamiliar, but because they do not align with its pre-existing, generalized concept of ``defective'', rendering the classification labels ineffective.

The standard method for adapting a pre-trained model to a new domain is supervised fine-tuning (SFT) with a labeled dataset~\citep{zhang2023instruction}. While direct SFT can provide modest improvements, its effectiveness is often limited by a deeper issue: a semantic gap between the complex video content and the sparse classification labels. High-level labels like ``pass'' or ``fail'' provide a weak supervisory signal, making it difficult for the model to learn the intricate visual patterns of the new domain. This forces the model into an inefficient learning scenario, where it must simultaneously master the visual grammar of a new environment (e.g., unique lighting, materials, and motion) and map it to a simple label. This often leads to overfitting on superficial correlations rather than a robust, causal understanding of the task, thereby limiting performance gains~\citep{hu2025sf2t, niu2025creft, han2024anchor, moenck2024industrial}.

To address these limitations, we introduce a novel rationale-bootstrapped two-stage fine-tuning framework. Our approach decomposes the difficult adaptation problem into two simpler, sequential steps. In the first stage, we employ a self-improvement mechanism inspired by Chain-of-Thought (CoT) prompting~\citep{wei2022chain}. We prompt the LVLM to generate detailed, step-by-step textual descriptions (or rationales) for unlabeled videos from the target domain. This creates a rich dataset of video-text pairs that we use for the initial fine-tuning stage, effectively teaching the model to ``speak the language'' of the target domain by grounding its visual understanding in descriptive text. This aligns with recent self-training methods where models learn from their own high-confidence, rationale-augmented outputs~\cite{huang2022large}. In the second stage, once the model has been equipped with a robust, domain-specific visual representation, we perform a standard SFT using the actual classification labels. By decoupling representation learning from classification, our framework enables the model to first build a strong foundation of the new domain's visual concepts before learning the final, high-level task. Our experiments demonstrate that this method significantly enhances classification performance in specialized domains.

The main contributions are threefold:

\begin{itemize}
    \item We first provide an in-depth analysis of the failure modes of direct SFT to domain adaptation for video classification, demonstrating how an intermediate rationale-generation stage effectively mitigates these issues.
    \item We propose the \textit{Rationale-Bootstrapped Two-Stage Fine-Tuning} (RB-FT) framework, a novel fine-tuning paradigm that effectively adapts LVLMs to new domains by first bridging the semantic gap with self-generated rationales, followed by task-specific tuning.
    \item We conduct a comprehensive empirical study on challenging video classification benchmarks, showing that RB-FT yields significant performance gains over zero-shot, direct SFT, and other strong baselines.
\end{itemize}

\section{Related Works}
\label{sec:related_works}
\subsection{Evolution of Video Classification}
The field of video classification has undergone a significant evolution. Early methods progressed from handcrafted spatio-temporal descriptors, e.g., 3D-SIFT~\citep{scovanner20073} and HOG3D~\citep{klaser2008spatio}, to trajectory-based pipelines such as Improved Dense Trajectories (iDT)~\citep{wang2013action}, which remained a strong pre-deep-learning baseline. Deep learning then supplanted manual feature engineering: two-stream CNNs separated appearance from motion, 3D CNNs (C3D) learned joint spatio-temporal filters~\citep{tran2015learning}, and CNN-RNN hybrids (e.g., LRCN) modeled temporal dependencies end-to-end~\citep{donahue2015long}. Subsequent architectures addressed long-range context and efficiency, from Temporal Segment Networks (TSN) to Inflated 3D ConvNets (I3D) and multi-rate designs like SlowFast~\citep{feichtenhofer2019slowfast}, with progress accelerated by large datasets such as Kinetics~\citep{carreira2017quo}.

\subsection{Transformer-based Models for Video Classification} 
Transformer architectures have fundamentally advanced video understanding by leveraging space-time self-attention to model long-range dependencies, often surpassing convolutional baselines~\citep{bertasius2021space, arnab2021vivit, liu2022video}. Concurrently, vision-language pretraining—pioneered by CLIP and ALIGN—introduced scalable natural-language supervision~\citep{radford2021learning, jia2021scaling}, which has been effectively adapted to video domains to enable zero-shot classification. More recently, the field has shifted toward multimodal instruction-tuned systems (e.g., BLIP-2, Video-LLaMA, and the GPT-4 series) that integrate visual encoders with LLMs to facilitate open-ended reasoning and audio-visual analysis~\citep{liu2023visual, li2023blip, awadalla2023openflamingo, achiam2023gpt, hurst2024gpt, comanici2025gemini}. Notably, approaches like LLaVA-CoT~\citep{xu2025llava} further enhance this paradigm by fine-tuning models to perform autonomous, multi-stage reasoning. Collectively, these advancements establish a robust foundation for prompt-controllable video classification with strong transfer capabilities.

\subsection{Self-Improvement Under Domain Shift}
While Vision Language Models (VLMs) exhibit strong generalization, they often degrade under significant distribution shifts, where abstract labels are insufficient to capture domain-specific semantics~\citep{kay2017kinetics}. Consequently, direct supervised fine-tuning (SFT) usually leads to overfitting or fails to bridge the semantic gap. To mitigate this, recent research has pivoted toward ``self-improvement'' strategies, utilizing Chain-of-Thought (CoT) prompting and model-generated rationales to provide dense supervisory signals~\citep{huang2022large, wei2022chain, wang2022self}. In particular, approaches such as distilling rationale from larger models~\citep{zhang2023video} and reflective self-training~\citep{cheng2024vision} demonstrate that incorporating intermediate reasoning steps significantly improves multimodal reasoning and transferability. Building on these insights, we propose Rationale-Bootstrapped Fine-Tuning (RB-FT). By adopting a two-stage paradigm: first aligning visual representations via detailed model-generated rationales, followed by label-based optimization, RB-FT effectively decouples representation learning from classification, offering a robust alternative to direct SFT in domain-shifted contexts.

\section{Method}
\label{sec:method}
\paragraph{Problem Formulation. }
The central problem addressed in this work is video classification under a significant domain shift. Let $M_\theta$ represent a pre-trained Vision-Language Model (VLM) with parameters $\theta$. We define the target domain dataset $\mathcal{D}_{target}$, which is partitioned into a training set $\mathcal{D}_{target}^{train}$ and a testing set $\mathcal{D}_{target}^{test}$:
\begin{equation*}
    \mathcal{D}_{target}^{train} = \{(v_i, y_i)\}_{i=1}^{N_{train}}, \quad \mathcal{D}_{target}^{test} = \{(v_j, y_j)\}_{j=1}^{N_{test}}
\end{equation*}
where $v$ represents a video instance and $y \in \{1, \dots, C\}$ denotes the corresponding class label. The distribution of $\mathcal{D}_{target}$ is assumed to differ substantially from the pre-training data of $M_\theta$.

Given a sample $\mathbf{s}=(V, T)$ consisting of a video $V$ and text $T$, the VLM processes each modality through specific encoders. The video input is temporally subsampled via $\mathcal{S}_{\tau}$ and spatially decomposed into patch tokens via $\Pi_{p,\ell}$, before being encoded by $\phi_{\mathrm{vid}}$. Concurrently, the text input is tokenized and embedded via $\phi_{\mathrm{text}}$. These multimodal tokens are augmented with positional embeddings ($\mathbf{p}$) and modality-type embeddings ($\mathbf{m}$), then concatenated into a unified sequence $\mathbf{X}$:
\begin{equation}
\begin{split}
\mathbf{X} = \mathrm{concat}\Big(
&\underbrace{\phi_{\mathrm{vid}}\circ\Pi_{p,\ell}\circ\mathcal{S}_{\tau}(V) + \mathbf{p}^{\mathrm{vid}} + \mathbf{m}^{\mathrm{vid}}}_{\text{Video Embeddings}}, \\
&\underbrace{E\circ\phi_{\mathrm{text}}(T) + \mathbf{p}^{\mathrm{text}} + \mathbf{m}^{\mathrm{text}}}_{\text{Text Embeddings}}
\Big)
\end{split}
\end{equation}
This sequence $\mathbf{X}$ is subsequently fed into the backbone Large Language Model (LLM). Our objective is to learn fine-tuned parameters $\theta^*$ using $\mathcal{D}_{target}^{train}$ that maximize classification accuracy on $\mathcal{D}^{test}_{target}$ while preserving the model's inherent multimodal reasoning capabilities.

\begin{figure*}[t]
  \centering
  \includegraphics[width=\textwidth]{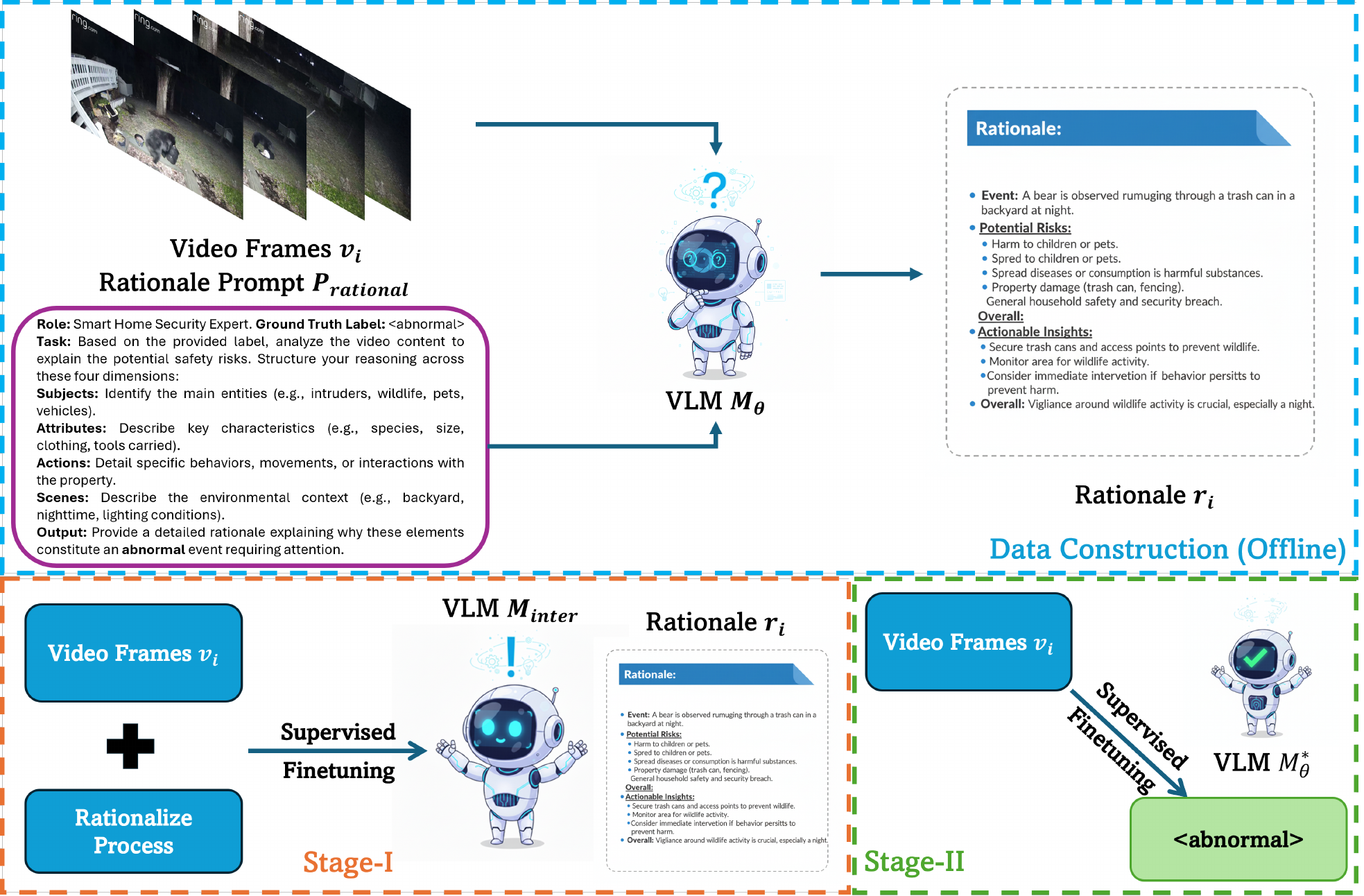}
  \caption{\textbf{Overview of the proposed Rationale-Bootstrapped Fine-Tuning Framework.} The pipeline consists of three phases: (Top) \textbf{Offline Data Construction}: A pre-trained VLM $M_\theta$ is leveraged to generate detailed textual rationales $r_i$ for training videos using a structured prompt $P_{rationale}$. This prompt conditions the model to adopt a specific expert persona (e.g., Smart Home Security Expert) and analyze the video across four semantic dimensions: subjects, attributes, actions, and scenes. (Bottom Left) \textbf{Stage-I (Rationale-Enhanced Self-Improvement)}: The model is supervised fine-tuned to generate these domain-specific rationales, producing an intermediate model $M_{inter}$ with enhanced reasoning capabilities. (Bottom Right) \textbf{Stage-II (Task-specific Label Alignment)}: The model undergoes a second stage of fine-tuning to predict the final ground-truth labels (e.g., \texttt{<abnormal>}), yielding the final optimized model $M^*_\theta$.}
  \label{fig:pipeline}
\end{figure*}

The proposed RB-FT framework achieves this by decomposing the adaptation process into two sequential stages. The first stage focuses on domain-semantic alignment through self-generated rationales, while the second stage focuses on task-specific classification. Both stages utilize the same training cohort. The overall pipeline is shown in~\Cref{fig:pipeline}.

\paragraph{Stage 1: Rationale-Enhanced Self-Improvement.}
The first stage of the RB-FT framework bridges the semantic gap between the generalist pre-trained model and the specialist target domain. This is achieved by creating and leveraging a dense, descriptive supervisory signal generated by the model itself, effectively guiding the model to learn the visual language of the new domain. This stage can be understood as a novel form of self-supervised pretext task, native to the capabilities of modern LVLMs. Traditional self-supervised learning (SSL) in video relies on pretext tasks like predicting clip order or temporal paces to force a model to learn meaningful representations without human labels. The goal of these tasks is to instill an understanding of intrinsic data properties, such as temporal coherence or motion dynamics. The rationale generation process serves a similar purpose but operates at a much higher semantic level. The pretext task becomes ``generating'' a coherent, detailed, and factually grounded textual description of this video.'' To succeed at this task, the model is compelled to learn to recognize the domain-specific objects, understand their complex temporal and spatial relationships, and accurately map these visual concepts to language. Unlike traditional SSL tasks that yield a single scalar loss, this pretext task generates a rich, structured textual output that serves as a powerful and dense supervisory signal for fine-tuning.

\myparagraph{Rationale Generation.} 
The framework initiates by leveraging the base VLM, $M_\theta$, to synthesize a textual rationale, $r_i$, for each video instance $v_i$ within the training subset $\mathcal{D}_{target}^{train}$. This is achieved through a structured prompting strategy that bridges the domain gap. We construct a specific prompt, $P_{rationale}$, which first conditions the model to adopt the persona of a ``Smart Home Security Expert,'' thereby aligning the generation with the specific safety-oriented requirements of the target domain.

To elicit high-quality, step-by-step reasoning, $P_{rationale}$ explicitly instructs the model to decompose its analysis across four semantic dimensions:
(1) \textbf{Subjects}: identifying primary entities such as wildlife, intruders, or vehicles;
(2) \textbf{Attributes}: detailing key characteristics like species, size, or clothing;
(3) \textbf{Actions}: capturing dynamic behaviors, movements, and interactions with the property; and
(4) \textbf{Scenes}: describing environmental contexts such as lighting conditions or backyard settings. The example is shown in~\Cref{fig:pipeline}.

Formally, for each video $v_i$, the rationale is generated via $r_i = M_{\theta}(v_i, P_{rationale})$. This offline process yields an intermediate rationale-augmented dataset, $\mathcal{D}_{rationale}^{train}=\{(v_i, r_i)\}_{i=1}^{N_{train}}$, which pairs raw visual inputs with rich, self-generated semantic supervision tailored to the target distribution.

\myparagraph{Intermediate Fine-Tuning.} With the rationale dataset, $D_{rationale}^{train}$ constructed, the next step is to perform an initial supervised fine-tuning of the base model $M_\theta$. The objective of this intermediate SFT is to align the model's visual representations more closely with the detailed, domain-specific textual descriptions which has just been generated. The model is trained to minimize the standard auto-regressive language modeling loss over the rationales, conditioned on the corresponding videos. The loss function for this stage, $\mathcal{L}_{\text{rationale}}$, is given by:

\[
\mathcal{L}_{\text{rationale}}
= - \sum_{i=1}^{N_{\text{train}}} \log P\!\left(r_i \mid v_i;\,\theta\right)
\]
$M_{inter}$ is obtained by minimizing the $\mathcal{L}_{\text{rationale}}$.

\paragraph{Stage 2: Task-specific Label Alignment.}

The second stage of the RB-FT framework fine-tunes the domain-adapted intermediate model, $M_{inter}$, for the final downstream classification task. This stage uses the original ground-truth dataset, $D_{target}^{train}=\{(v_i, y_i)\}^{N_{train}}_{i=1}$. 

A second round of SFT is performed, starting from the weights of $M_{inter}$. The objective is to minimize the standard cross-entropy loss for classification. The prompt for this stage is a simple classification query. The loss function, $\mathcal{L}_{\text{classify}}$, is:

\[
\mathcal{L}_{\text{classify}}
= - \sum_{i=1}^{N_{\text{train}}} \log P\!\left(y_i \mid v_i;\,\theta_{inter}\right)
\]

This fine-tuning stage is hypothesized to be significantly more effective than direct SFT on the base model $M_{\theta}$. Because $M_{inter}$ has already developed a feature space that is well adapted to the visual nuances of the target domain, the optimization landscape for the classification task is much more favorable. The model is no longer burdened with the dual challenge of learning the domain's visual language and the classification task simultaneously. Instead, the final SFT stage primarily learns to map the already meaningful and domain-aligned features to the discrete set of class labels. This separation of concerns substantially reduces the risk of catastrophic forgetting of the model's core reasoning abilities and mitigates the tendency to overfit to spurious correlations in the limited labeled data. The final output of this stage is the fully adapted model, $M_\theta^*$.


\section{Experiments}
\label{sec:experiments}
We evaluate our proposed RB-FT framework on two domain-specific video classification benchmarks: SmartHome-LLM (abnormal vs.~normal daily activities) and MultiHateClip (hateful vs.~normal video memes). We adopt Qwen2-VL-7B-Instruct~\citep{wang2024qwen2} and Qwen2.5-VL-7B-Instruct~\citep{bai2025qwen2} as our backbone VLMs.
We focus on the following three research questions, which collectively examine the central hypothesis of our paper, that bootstrapping domain-specific rationale understanding provides a strictly more transferable and more semantically stable representation prior to domain SFT:\\
\textbf{RQ1: Quantitative Effect.} Does RB-FT consistently outperform direct supervised fine-tuning (Direct-SFT) across different backbones and datasets under the same training budget?\\
\textbf{RQ2: Factorization and Ablation.} Which component(s) of RB-FT contribute the most to the final gains? In particular: (1) Where should the rationale be placed in the prompt? and (2) Does mixing self-generated rationales benefit more than partially using ground-truth rationales?\\
\textbf{RQ3: Causal Interpretability.} Does RB-FT change ``what the model looks at''? For example, are attention maps qualitatively and quantitatively more causally grounded (e.g., more sensitive to masking of semantically key objects) than Direct-SFT?

\subsection{Experimental Setup}

\myparagraph{Datasets.} We evaluate our framework on two domain-specific video classification datasets.
\noindent \textbf{SmartHome-LLM Benchmark}~\citep{zhao2025smarthome} is a video anomaly detection benchmark featuring $1,203$ smart home clips across seven categories (e.g., Wildlife, Senior Care). Annotations include anomaly tags, detailed descriptions, and reasoning. The dataset's primary challenges lie in the subtlety of anomalies, high contextual diversity, and the requisite common-sense reasoning to distinguish normal from abnormal events.
\noindent \textbf{MultiHateClip}~\citep{wang2024multihateclip} is a multilingual benchmark for hateful video detection, comprising $2,000$ videos annotated as hateful, offensive, or normal. We utilize the English-language subset for our experiments. The benchmark's key challenges include the need for cultural nuance, the inherently multimodal nature of hate speech (encompassing visual, audio, and textual elements), and the fine-grained distinction between ``hateful'' and ``offensive'' content.

\myparagraph{Implementation Details.} All experiments are conducted on $8$ NVIDIA H100 GPUs. We implement the RB-FT pipeline using the Qwen2-VL-7B-Instruct~\citep{wang2024qwen2} and Qwen2.5-VL-7B-Instruct~\citep{bai2025qwen2} models as our base VLMs. Videos are sampled at $1$ FPS with a maximum resolution of $360\times420$ pixels. For both fine-tuning stages, we use a cosine learning rate scheduler with a $3\%$ warm-up rate and a weight decay of $0.1$. The learning rates are set to $1\times10^{-5}$ for the language model and merger components and $2\times10^{-6}$ for the vision tower. We enable gradient checkpointing and clip gradients at a norm of $1.0$. Training is performed with a global batch size of 16. Both Stage-I and Stage-II are trained for a single epoch to mitigate the risk of overfitting on the specialized datasets.

\myparagraph{Baselines and Metrics.} To establish a robust performance benchmark, we compare the proposed RB-FT framework with several baselines to rigorously validate its effectiveness. 
Specialized Video Models: UniFormerV2~\citep{li2022uniformerv2} and InternVideo2~\citep{wang2024internvideo2}, which represent the state-of-the-art in models designed specifically for video understanding tasks.
General-Purpose VLMs: GPT-4o-mini, GPT-4o~\citep{hurst2024gpt}, Gemini-2.5-Flash/Pro~\citep{comanici2025gemini}, which are powerful, proprietary large multimodal models.
We also compare against the zero-shot performance of our base models and the standard Direct-SFT approach.

Performance is evaluated using \textbf{Accuracy (Acc)} for overall classification performance and \textbf{F1-score} to assess the balance between precision and recall for each class, which is particularly important for datasets with imbalanced classes, such as MultiHateClip.

\subsection{Main Quantitative Results}
\label{subsec:quantitative_results}
The quantitative results, presented in~\Cref{tab:smarthome} and~\Cref{tab:multihateclip}, demonstrate the consistent and significant effectiveness of the RB-FT framework.

On the SmartHome-LLM dataset (\Cref{tab:smarthome}), RB-FT applied to Qwen2.5-VL-7B achieves 82.65\% accuracy, outperforming Direct-SFT by a substantial 6.63 percentage points and the zero-shot baseline by 27.55 percentage points. The improvement is particularly pronounced in the F1-score for the ``Normal'' class, which jumps from 51.55\% with Direct-SFT to 76.06\% with RB-FT. This indicates that the rationale-bootstrapped alignment stage helps the model develop a much stronger and more reliable understanding of what constitutes normalcy within the smart home domain, a key challenge in anomaly detection. Our method also surpasses all other baselines, including the powerful Gemini-2.5-Pro (73.47\%).

On the MultiHateClip dataset (\Cref{tab:multihateclip}), RB-FT with Qwen2.5-VL-7B attains 71.00\% accuracy, a 4.14 percentage point improvement over Direct-SFT. Critically, RB-FT demonstrates enhanced learning for underrepresented categories. The F1-score for the minority ``Hateful'' class more than doubles, improving from 11.11\% to 23.53\%. This shows that the dense supervisory signal from the rationales enables the model to learn more discriminative features for classes with fewer examples, which are often overlooked during standard SFT. While Gemini-2.5-Pro achieves a higher overall accuracy (75.81\%), it completely fails on the ``Offensive'' class (0.00\% F1), highlighting its lack of robustness. In contrast, RB-FT delivers a much more balanced and reliable performance across all three classes.

The consistent pattern of improvement across both the Qwen2-VL-7B and Qwen2.5-VL-7B models validates the robustness of the RB-FT framework, suggesting that its benefits in addressing fundamental domain adaptation challenges are independent of minor architectural variations.
\begin{table*}[h]
\centering
\caption{Quantitative results on the SmartHome-LLM benchmark.}
\scriptsize
\label{tab:smarthome}
\setlength{\tabcolsep}{9pt}
\begin{tabular}{@{}lllrrr@{}}
\toprule
Dataset & Model & Method & Accuracy (\%) $\uparrow$ & F1 (Normal) $\uparrow$ & F1 (Abnormal) $\uparrow$ \\
\midrule
\multirow{12}{*}{SmartHome-LLM}
  & \multirow{6}{*}{---}
    & UniFormerV2       & 70.12 & 38.02 & 78.95 \\
  &                    & InternVideo2      & 68.70 & 35.40 & 79.10 \\
  &                    & GPT-4.1           & 70.53 & 69.23 & 71.72 \\
  &                    & Gemini-2.5-Flash  & 73.33 & 70.79 & 75.47 \\
  &                    & GPT-4o            & 62.76 & 63.32 & 62.18 \\
  &                    & Gemini-2.5-Pro    & 73.47 & 70.11 & 76.15 \\
\cmidrule(lr){2-6} 
  & \multirow{3}{*}{Qwen2-VL-7B}
    & Zero-shot        & 57.65 & 59.11 & 56.08 \\
  &                    & Direct-SFT       & 69.39 & 36.17 & 79.87 \\
  &                    & RB-FT            & 80.61 & 74.67 & 84.30 \\
\midrule
  & \multirow{3}{*}{Qwen2.5-VL-7B}
    & Zero-shot        & 55.10 & 60.36 & 48.24 \\
  &                    & Direct-SFT       & 76.02 & 51.55 & 84.07 \\
  &                    & RB-FT            & \textbf{82.65} & \textbf{76.06} & \textbf{86.40} \\
\bottomrule
\end{tabular}
\end{table*}

\begin{table*}[h]
\centering
\caption{Quantitative results on the MultiHateClip benchmark.}
\scriptsize
\label{tab:multihateclip}
\setlength{\tabcolsep}{9pt}
\begin{tabular}{@{}lllrrrr@{}}
\toprule
Dataset & Model & Method & Accuracy (\%) $\uparrow$ & F1 (Normal) $\uparrow$ & F1 (Hateful) $\uparrow$ & F1 (Offensive) $\uparrow$ \\
\midrule
\multirow{10}{*}{MultiHateClip}
  & \multirow{4}{*}{---}
    & UniFormerV2       & 59.17 & 72.10 & 8.33  & 35.42 \\
  &                    & InternVideo2      & 62.72 & 74.85 & 10.91 & 38.06 \\
  &                    & GPT-4.1                    & 70.46 & 83.58 & 60.38 & 23.53  \\
  &                    & Gemini-2.5-Flash           & 70.94 & 83.98 & 18.18 & 40.00  \\ 
  &                    & GPT-4o                     & 72.19 & 84.16 & 55.45 & 12.50  \\
  &                    & Gemini-2.5-Pro             & 75.81 & 84.71 & 61.11 & 0.00  \\ 
\cmidrule{2-7} 
  & \multirow{3}{*}{Qwen2-VL-7B}
    & Zero-shot         & 53.25 & 65.33 & 0.00  & 39.37 \\
  &                    & Direct-SFT        & 65.68 & 78.81 & 4.76  & 29.63 \\
  &                    & RB-FT             & 68.64 & 81.27 & 14.29 & 42.65 \\
\cmidrule{2-7} 
  & \multirow{3}{*}{Qwen2.5-VL-7B}
    & Zero-shot         & 54.48 & 66.90 & 2.70  & 40.12 \\
  &                    & Direct-SFT        & 66.86 & 79.92 & 11.11 & 41.35 \\
  &                    & RB-FT             & \textbf{71.00} & \textbf{83.47} & \textbf{23.53} & \textbf{49.78} \\
\bottomrule
\end{tabular}
\end{table*}

\subsection{In-Depth Analysis and Discussion}
\label{subsec:ablation}
We conduct a series of ablation studies to understand the mechanisms behind RB-FT's effectiveness. These studies collectively reveal a coherent narrative about how the framework operates: the specific composition of the rationale supervision enables a scalable self-generation process, which results in a more interpretable and robust final model. All the ablation studies are conducted on the SmartHome-LLM dataset.

\myparagraph{Ablation Study: Impact of Rationale Composition.}
We examine how the composition of the rationales affects the learning. The term rational is fixed and refers only to the generated reasoning description. We compare three supervision formats that differ only in whether and where the final classification appears relative to the rationale, while keeping the content of the rationale unchanged. We denote P as the input prompt, R as the rational, and C as the final class label. The formats are P+R with no explicit class label, P+C+R with the class label placed before the rational, and P+R+C with the class label placed after the rational. During fine-tuning, the model is trained to produce the target in the specified order, and at evaluation, we elicit outputs in the same order.

Across our experiments, the P+R setting achieves the best overall performance, surpassing both variants with an explicit class label. The gains arise because supervision centered on the rationales focuses learning on the reasoning steps rather than reproducing label tokens. Omitting the label discourages shortcut behavior, reduces label leakage, and encourages the model to ground its prediction in event descriptions, object interactions, and causal relations. Placing the label at the beginning promotes post hoc justification, while putting it at the end still introduces a strong cue that can be memorized. Training that concentrates on the rationales alone yields a denser and more informative signal and improves generalization, calibration, and robustness for tasks that require multi-step reasoning.

\begin{table*}[ht]
\centering
\scriptsize
\setlength{\tabcolsep}{3pt}

\begin{minipage}[t]{0.48\textwidth}
\centering
\caption{Ablation Study on the Composition of the Rationales.}
\label{tab:composition}
\begin{tabular}{@{}llrrr@{}}
\toprule
Model & Rationale & Acc (\%) $\uparrow$ & F1 (Normal) $\uparrow$ & F1 (Abnormal) $\uparrow$ \\
\midrule
\multirow{3}{*}{Qwen2-7B-VL}   & P+C+R & 79.59 & 73.33 & 83.47 \\
                               & P+R+C & 80.61 & 73.97 & 84.55 \\
                               & P+R   & 80.61 & 74.67 & 84.30 \\
\midrule
\multirow{3}{*}{Qwen2.5-7B-VL} & P+C+R & 81.10 & 74.90 & 85.30 \\
                               & P+R+C & 81.65 & 75.10 & 85.70 \\
                               & P+R   & \textbf{82.65} & \textbf{76.06} & \textbf{86.40} \\
\bottomrule
\end{tabular}
\end{minipage}
\hfill
\begin{minipage}[t]{0.48\textwidth}
\centering
\caption{Ablation Study on the Ratio of Self-Generated Rationales in Stage-I.}
\label{tab:ratio}
\begin{tabular}{@{}llrrr@{}}
\toprule
Model & Ratio & Acc (\%) $\uparrow$ & F1 (Normal) $\uparrow$ & F1 (Abnormal) $\uparrow$ \\
\midrule
\multirow{3}{*}{Qwen2-7B-VL}   & 20\%  & 71.43 & 53.20 & 81.90 \\
                               & 60\%  & 74.49 & 56.80 & 82.70 \\
                               & 100\% & \textbf{80.61} & \textbf{74.67} & \textbf{84.30} \\
\midrule
\multirow{3}{*}{Qwen2.5-7B-VL} & 20\%  & 76.53 & 59.80 & 85.40 \\
                               & 60\%  & 78.57 & 62.10 & 85.90 \\
                               & 100\% & \textbf{82.65} & \textbf{76.06} & \textbf{86.40} \\
\bottomrule
\end{tabular}
\end{minipage}

\end{table*}

\myparagraph{Ablation Study: Ratio of Self-Generated Rationales in Stage-I.}
We next evaluate the impact of using self-generated rationales versus human-annotated ones by training models with varying ratios of synthetic data in Stage-I.~\Cref{tab:ratio} shows a clear and positive trend: performance improves as the proportion of self-generated rationales increases, with the 100\% self-generated setting yielding the best results.

This result strongly validates the premise that the model's own descriptions, even if imperfect, are a powerful and effective source of supervision. It suggests that for this domain adaptation task, the breadth of descriptive coverage across the entire training set (achieved with 100\% self-generated data) is more valuable than the potentially higher quality of a small subset of human annotations. This has profound implications for annotation efficiency, proving that the rationale-based learning signal is robust enough to be self-generated at scale. The goal is not to model the distribution of rationales but to use them as a rich signal to learn better visual representations.

\begin{table*}[!t]
\centering
\caption{Ablation Study on the Key Objects.}
\label{tab:key_objects}
\scriptsize
\setlength{\tabcolsep}{9pt}
\begin{tabular}{@{}cccccc@{}}
\toprule
Model & Paradigm & Input & Acc. (\%) $\uparrow$ & F1 (Normal) $\uparrow$ & F1 (Abnormal) $\uparrow$ \\
\midrule
\multirow{6}{*}{Qwen2-VL-7B} & \multirow{3}{*}{Direct\_SFT} & Rand. Masked & 66.84 & 21.69 & 78.96 \\
                             &                              & Obj. Masked  & 67.86 & 18.18 & 80.00 \\
                             &                              & Ori. Frames  & 69.39 & 36.17 & 79.87 \\
\cmidrule{2-6}
                             & \multirow{3}{*}{RB-FT}       & Rand. Masked & 75.51 & 63.08 & 81.68 \\
                             &                              & Obj. Masked  & 66.33 & 40.00 & 76.60 \\
                             &                              & Ori. Frames  & \textbf{80.61} & \textbf{74.67} & \textbf{84.30} \\
\bottomrule
\end{tabular}
\end{table*}

\begin{figure*}[!t]
  \centering
  \includegraphics[width=\textwidth]{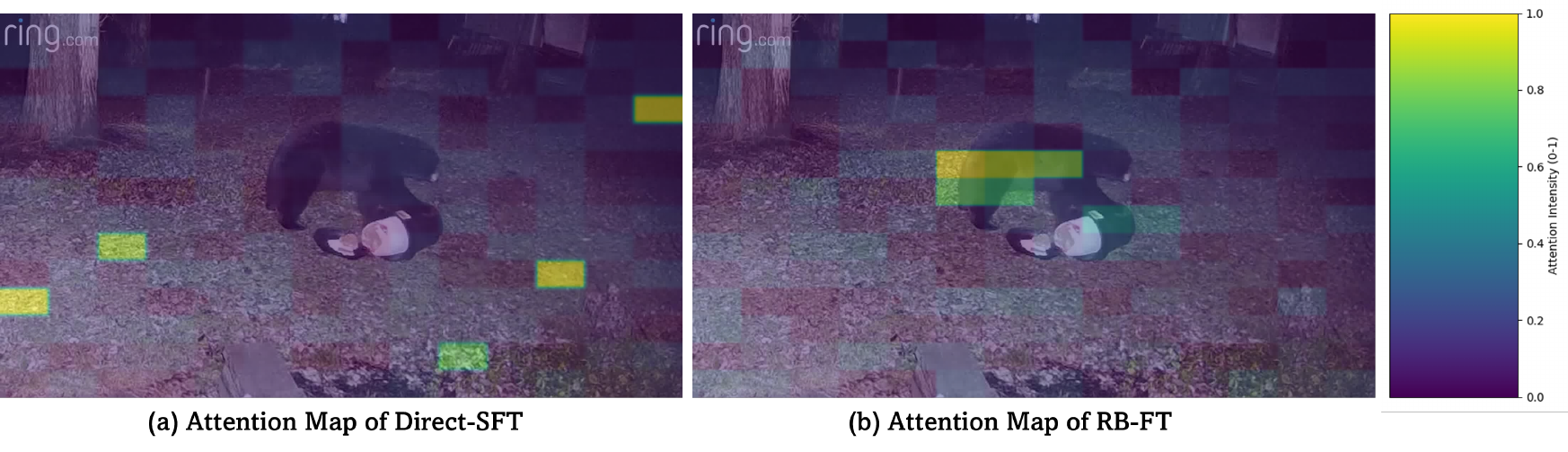}
  \caption{Comparative visualization of attention maps between the baseline Direct-SFT model (a) and our proposed RB-FT model (b). The heatmaps (purple to yellow) represent attention intensity, scaled from 0 to 1. The RB-FT model (b) demonstrates significantly improved focal accuracy, concentrating high-intensity attention on the salient subjects (the black bear and the trash can), whereas the Direct-SFT model (a) exhibits diffuse attention, failing to localize the critical regions of interest.}
  \label{fig:attn_analysis}
\end{figure*}

\myparagraph{Ablation Study: Impact of Key Objects.}
To investigate whether RB-FT learns a more causal and interpretable model, we perform an object masking ablation. We identify the three most salient objects in each test video and compare the model's performance on the original frames, frames with these objects masked, and frames with random patches of equivalent area masked.

The results are shown in~\Cref{tab:key_objects}. For the RB-FT model, masking critical objects (Obj. Masked) causes a dramatic drop in accuracy (from 80.61\% to 66.33\%). This drop is far more significant than that caused by masking random patches (Rand. Masked, 75.51\%) and substantially larger than the drop observed for the Direct-SFT model (69.39\% to 67.86\%). This provides strong evidence that the RB-FT model has learned to ground its classifications in semantically meaningful objects. These very objects would be explicitly named in the generated rationales. In contrast, the Direct-SFT model relies more on diffuse, superficial correlations across the entire frame, making it less sensitive to removing specific objects. This demonstrates that the rationale-based supervision in Stage 1 produces a model that is more accurate, more interpretable, and causally aligned with the task.

\subsection{Attention Mode Analysis}
\label{subsec:attn_analysis}
To qualitatively assess the impact of our proposed RB-FT, we visualized the attention distributions of the baseline Direct-SFT model and our RB-FT model on a representative sample, shown in~\Cref{fig:attn_analysis}. The resulting attention maps illustrate the models' ability to localize key objects within the input. As evidenced in the comparison, the Direct-SFT approach (a) produces a scattered and unfocused attention pattern, erroneously assigning importance to irrelevant background elements while failing to identify the critical subjects. In stark contrast, our proposed RB-FT method (b) demonstrates exceptional effectiveness, generating a highly concentrated attention map that precisely localizes the key entities in the scene—namely, the person on the ground and the bear. This visualization confirms that the RB-FT technique successfully guides the model's focus to semantically critical regions, suppressing background noise and enabling a more robust, accurate understanding of the scene's content.

\section{Conclusion}
\label{sec:conclusion}
This work addresses the challenge of adapting Large Vision Language Models (LVLMs) to specialized video domains characterized by pronounced semantic gaps, where direct supervised fine-tuning often proves inadequate. To bridge this disconnect, we introduce Rationale-Bootstrapped Fine-Tuning (RB-FT), a framework that decouples domain-semantic alignment from task classification. By leveraging self-generated rationales as dense intermediate supervision, RB-FT effectively aligns visual representations with domain-specific concepts before label optimization. Extensive experiments demonstrate that this approach significantly outperforms standard baselines, yielding robust, interpretable representations.

\bibliography{custom}

\appendix



\end{document}